\documentclass{article} 
\usepackage{iclr2027_conference,times}

\iclrfinalcopy

\usepackage{amsmath,amsfonts,bm}

\def\eqref#1{equation~\ref{#1}}

\def\1{\bm{1}}

\DeclareMathAlphabet{\mathsfit}{\encodingdefault}{\sfdefault}{m}{sl}
\SetMathAlphabet{\mathsfit}{bold}{\encodingdefault}{\sfdefault}{bx}{n}

\usepackage{hyperref}
\usepackage{url}
\usepackage{booktabs}
\usepackage{graphicx}
\usepackage{multirow}
\usepackage{enumitem}
\newcommand{\PAR}[1]{\vskip1pt \noindent{\bf #1}}

\usepackage{marvosym}

\usepackage{pifont}
\usepackage{caption}
\usepackage{subcaption}
\usepackage{wrapfig}
\usepackage{titletoc}
\usepackage{tikz}
\usepackage{float}
\newcommand{\cmark}{\ding{51}}
\newcommand{\xmark}{\ding{55}}

\usepackage[svgnames]{xcolor}
\definecolor{mylightblue}{RGB}{85, 137, 201}

\hypersetup{
    colorlinks=true,
    citecolor=mylightblue,
    linkcolor=mylightblue,
    urlcolor=mylightblue
}

\title{MORE: Learning Motion-Oriented Registers for Driving World Models and Planning}

\newcommand{\authorskipshort}{\hspace{6mm}}
\author{
\begin{tabular}{l}
Yingdong Gu$^{1}$\thanks{Equal contribution, $\textsuperscript{\Letter}$Corresponding author.} \authorskipshort
Haoyu Wang$^{2*}$ \authorskipshort
Shaocheng Yan$^{1}$ \authorskipshort
Zhenjun Zhao$^{3}$ \authorskipshort \\
Hao Han$^{2}$ \authorskipshort 
Pengcheng Shi$^{1}$ \authorskipshort
Xianfei Li$^{2}$ \authorskipshort
Qinghua Tan$^{1}$ \authorskipshort \\
Wenlong Liao$^{2}$ \authorskipshort
Tao He$^{2}$ \authorskipshort
Jiayuan Li$^{1}\textsuperscript{\Letter}$\authorskipshort
Pai Peng$^{2}\textsuperscript{\Letter}$ \\
\end{tabular}
\\[3mm]
\begin{tabular}{l}
$^1$Wuhan University \authorskipshort
$^2$CowaRobot Co. Ltd. \authorskipshort
$^3$University of Zaragoza  \\
\hspace{0.5mm} {\texttt{\small \{guyingdong, ljy\_whu\_2012\}@whu.edu.cn} }
\\[1mm]
\hspace{1mm} Code: \small{\url{https://github.com/Cyril-gyd/MORE}}
\end{tabular}
}

\begin{document}

\maketitle

\fancyhf{}
\fancyfoot[C]{\thepage}
\renewcommand{\headrulewidth}{0pt}

\begin{abstract}

By predicting future states, latent world models provide rich self-supervision for end-to-end driving, but often inherit dense features from pretrained visual encoders as their world-state interface. This creates a mismatch: visual pretraining produces information-rich, spatially dense features, but their adaptation for motion-conditioned world modeling and planning is deferred to downstream training. The dense grid also makes future prediction more computationally expensive as visual resolution increases. We therefore argue that video pretraining should directly learn state representations suited to downstream world modeling and planning, rather than relying on downstream training to adapt inherited visual features. We propose \textbf{\textit{MORE}}, a V-JEPA framework that learns \textbf{M}otion-\textbf{O}riented \textbf{RE}gisters during video pretraining as a shared state interface for driving world models and planning. MORE injects temporally aligned ego motion into a small set of registers and uses them as the sole context for masked patch prediction, channeling dense patch-level supervision through a compact state bottleneck. A complementary blockwise register-masking objective further shapes predictive structure in the register space. During downstream training, the same register space defines the future states predicted by the world model and provides shared scene memory for trajectory generation and scoring. On NAVSIM-v2, MORE outperforms the state of the art by 1.6 in EPDMS, while achieving superior efficiency with only 1/10 the inference latency of DriveLaW. These results support learning the state interface for world modeling and planning directly during video pretraining.

\end{abstract}

\section{Introduction}
\label{sec:introduction}

End-to-end autonomous driving~\citep{wu2022trajectory,hu2023planning,vad,sparsedrive} predicts future trajectories from sensor observations, but low-dimensional expert trajectories provide limited supervision for high-capacity visual encoders. Driving world models provide richer signals by predicting future observations~\citep{wang2024driving,li2026drivevla,wang2024drivedreamer}, occupancy or BEV states~\citep{zheng2024occworld,yang2025driving,resworld}, or latent features~\citep{li2025enhancing,zheng2025world4drive}.  Joint-Embedding Predictive Architectures (JEPA)~\citep{jepa} offer a complementary route by learning predictable representations rather than reconstructing pixels. V-JEPA~\citep{VJEPA} extends this principle to spatiotemporal masked prediction in videos, while V-JEPA 2-AC~\citep{V-jepa2} and Drive-JEPA~\citep{drive-jepa} further demonstrate its potential for action-conditioned prediction and trajectory planning.

\begin{figure}[ht]
\begin{centering}
\includegraphics[scale=0.173]{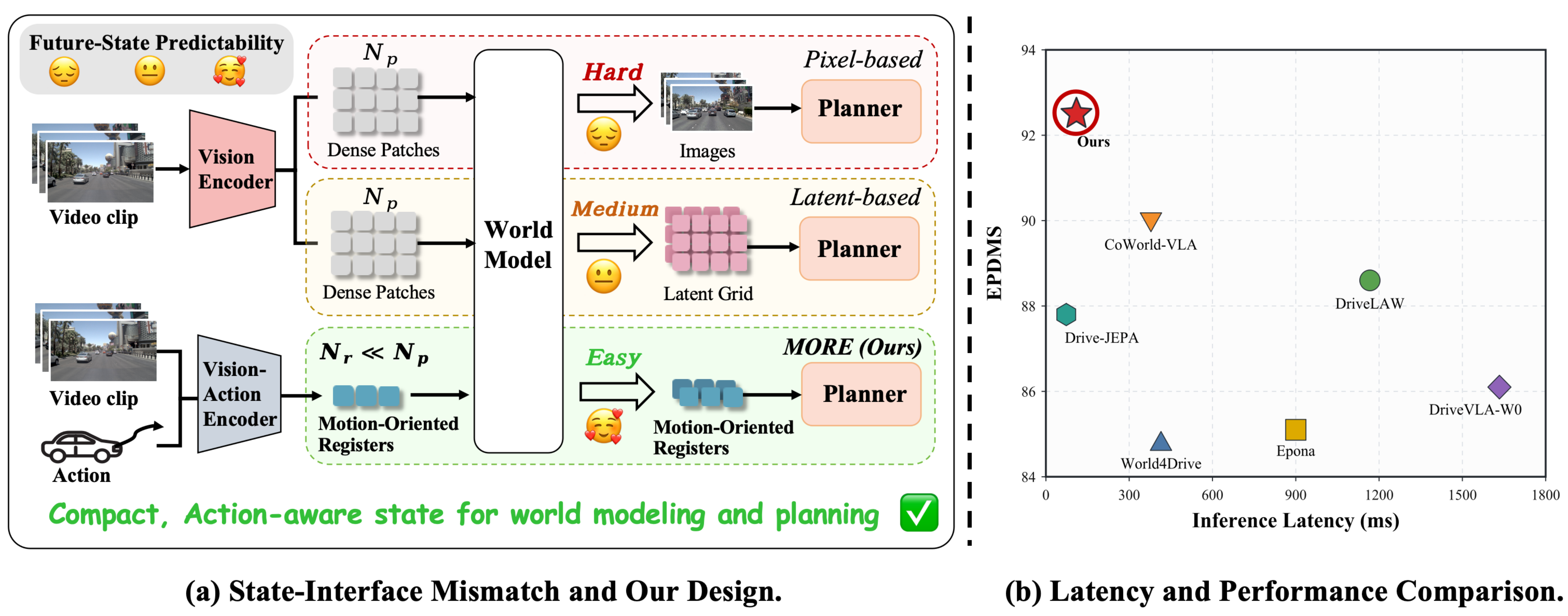}
\end{centering}
\captionsetup{skip=6pt}
\caption{\textbf{State-interface mismatch and inference efficiency.}
(a) Comparison of dense state interfaces and MORE's compact motion-oriented register interface for world modeling and planning.
(b) EPDMS versus inference latency.}
\label{Fig1}
\vspace{-10pt}
\end{figure}

However, a fundamental mismatch remains between pretrained visual representations and the states required by downstream world models and planners. Visual pretraining primarily learns information-rich representations of scene content and spatiotemporal structure. In contrast, a world model requires states that capture scene evolution conditioned on ego motion and support future prediction~\citep{driveworld}, while a planner further requires decision-relevant context. Existing approaches typically learn visual representations first and adapt them during downstream training~\citep{worldrft,zheng2025world4drive,li2025enhancing}. For example, V-JEPA 2-AC~\citep{V-jepa2} learns action-conditioned dynamics over frozen visual features after action-free pretraining. LAW~\citep{li2025enhancing} introduces ego-trajectory-conditioned future latent prediction during end-to-end training, while WorldRFT~\citep{worldrft} further aligns scene representations with planning through geometric priors and planning supervision. Together, these methods indicate that rich visual features do not automatically form ego-motion-sensitive world states suited to future prediction, leaving much of the structure for state evolution and control to be learned downstream. We refer to this gap, illustrated in Fig.~\ref{Fig1}(a), as a \textbf{\emph{state-interface mismatch}}.

Beyond this alignment gap, the organization of future states poses a related efficiency challenge. V-JEPA 2-AC~\citep{V-jepa2} predicts spatiotemporally indexed patch features, DriveX~\citep{drivex} and WoTE~\citep{WoTE} predict latent BEV grids, and World4Drive~\citep{zheng2025world4drive} predicts spatial visual latents. Because these future states retain spatial grids, finer grids require more tokens to be predicted at each future step. DrivoR~\citep{drivor} and NTR~\citep{ntr} instead compress dense visual features into a few planning tokens. This reduces state size for current-scene planning, but compactness alone does not ensure ego-motion sensitivity or the predictive structure needed for future modeling.

Our central question is therefore not how to further compress visual features during downstream training, but whether video pretraining can directly learn state representations suited to subsequent world modeling and planning. The key challenge is to retain dense visual supervision while endowing this state space with sufficient information, ego-motion sensitivity, and predictive structure.

To instantiate this principle, we propose \textit{MORE}, a V-JEPA framework that directly learns motion-oriented register states. During video pretraining, we introduce a small set of registers into the video encoder and condition them on temporally aligned ego motion. To ensure that these registers retain sufficient visual information, we require masked patch representations to be predicted solely through the registers. Dense patch-level supervision is therefore routed through a compact register bottleneck and directly shapes the state interface used downstream. We further apply blockwise masking and prediction in the register space to explicitly encourage predictive state structure. During joint training, we retain the same register space. A non-autoregressive world model predicts future registers from observed registers and ego motion, while the planner uses both as shared scene memory for trajectory generation and scoring. The world model can therefore model future evolution directly in the state space shaped during pretraining, rather than constructing predictive states from generic visual features. By processing only a small set of registers, MORE also reduces downstream computation, yielding the low inference latency shown in Fig.~\ref{Fig1}(b) and detailed in Appendix~\ref{efficiency}.

Our contributions are summarized as follows:
\begin{itemize}[leftmargin=*, itemsep=2pt, topsep=1pt]
    \item We identify and address the \textbf{\emph{state-interface mismatch}} between pretrained visual representations and the states needed for downstream world modeling and planning by learning a shared state interface during video pretraining.

    \item We introduce motion-oriented registers, register-mediated masked patch prediction, and blockwise register masking to jointly promote information sufficiency, motion sensitivity, and predictive structure in the learned state space.
    
    \item We use the same register state space for non-autoregressive world modeling and end-to-end trajectory planning, and evaluate the resulting framework on NAVSIM and NAVSIM-v2, together with closed-loop driving evaluation in HUGSIM.
\end{itemize}

\section{Related Works}
\label{sec:related_work}

\paragraph{Joint-Embedding Predictive Architectures and World Modeling.}
JEPA~\citep{jepa} shifts prediction from observation space to learned representation space. I-JEPA~\citep{I-JEPA} predicts masked image-region representations from visible context, V-JEPA~\citep{VJEPA} extends this principle to video, and V-JEPA 2~\citep{V-jepa2} scales both model size and pretraining data. V-JEPA 2.1~\citep{vjepa2.1} further improves fine-grained spatial and temporal representations through dense predictive loss and deep self-supervision. Recent work explores JEPA for action-conditioned latent world modeling. V-JEPA 2-AC~\citep{V-jepa2} freezes a pretrained video encoder and predicts future dense patch features conditioned on actions, while systematic studies of JEPA-WMs examine multi-step latent prediction, action conditioning, and planning strategies~\citep{terver2025drives}. For autonomous driving, Drive-JEPA~\citep{drive-jepa} transfers V-JEPA representations to trajectory learning. These works demonstrate the potential of JEPA for prediction and control, but the downstream state is generally inherited from visual pretraining, with action-conditioned state evolution introduced mainly during post-training. We instead shape the state space used for future modeling and planning directly during video pretraining.

\PAR{World Models for Autonomous Driving.}
Autonomous-driving world models predict the future in observation space, structured spatial representations, or latent feature space. DriveDreamer~\citep{wang2024drivedreamer} and Drive-WM~\citep{wang2024driving} generate action- or trajectory-conditioned images and videos; DriveVLA-W0~\citep{li2026drivevla} uses future image prediction as dense self-supervision.
These approaches preserve rich visual information but also model appearance details not directly relevant to planning. OccWorld~\citep{zheng2024occworld} and Drive-OccWorld~\citep{yang2025driving} instead predict occupancy states, DriveWorld~\citep{driveworld} and DriveX~\citep{drivex} model dynamics in structured scene representations, and WoTE~\citep{WoTE} and ResWorld~\citep{resworld} use future BEV states to evaluate or refine trajectories. Such representations provide explicit geometry but require a full spatial grid at every future step.

Latent world models avoid explicit observation reconstruction by predicting in feature space~\citep{li2025enhancing,dawn, zheng2025world4drive,rise}. LAW~\citep{li2025enhancing} uses ego-trajectory-conditioned future latent prediction as an auxiliary objective, World4Drive~\citep{zheng2025world4drive} predicts future visual latents for trajectory selection, and WorldRFT~\citep{worldrft} aligns latent representations with planning through geometric and decision supervision. These methods validate latent prediction for driving, but their states are typically derived from existing visual representations and adapted to dynamics and planning through downstream objectives. Dense latent grids also keep state size tied to visual resolution. We instead pretrain the downstream state space itself, using compact registers as both future prediction targets and planning context.

\section{Method}
\label{sec:method}

\begin{figure}[ht]
\begin{center}
\includegraphics[width=\linewidth]{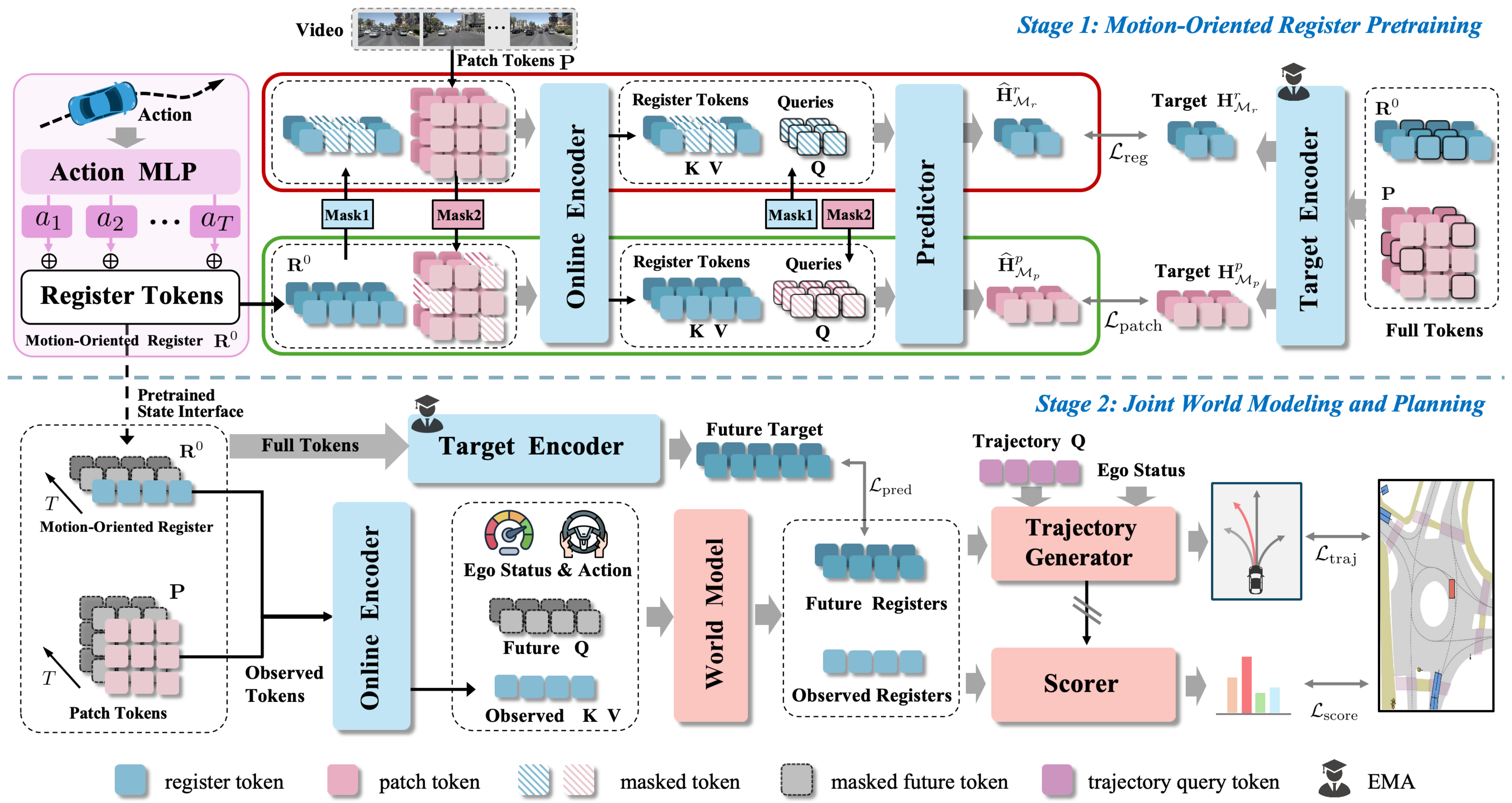}
\end{center}
\captionsetup{skip=1pt}
\caption{\textbf{Overview of \textit{MORE}.} During pre-training, motion-oriented registers are learned through register-mediated masked patch prediction and blockwise register masking. During joint training, the same register space is used for world modeling and as shared scene memory for trajectory generation and scoring.
}
\label{pipeline}
\vspace{-10pt}
\end{figure}

We propose \textit{MORE}, which learns motion-oriented registers as a compact state interface for driving world models and planning. As shown in Fig.~\ref{pipeline}, our pipeline consists of two stages. During pretraining, ego-motion conditioning, register-mediated patch prediction, and blockwise register masking shape an informative and predictable register state space from dense supervision. During joint training, we reuse the same register space to train the world model, trajectory generator, and scorer.

\subsection{Preliminaries on Joint-Embedding Predictive Architectures}

Joint-Embedding Predictive Architectures (JEPA)~\citep{jepa} learn representations by predicting latent targets rather than reconstructing observations. In the asymmetric online--target formulation, an online encoder processes a corrupted view, while a target encoder updated by the exponential moving average (EMA) provides targets from the corresponding clean view.

V-JEPA~\citep{VJEPA} applies this principle to masked video representation prediction. Let $\mathbf{y}$ denote the complete video, $\mathcal{M}$ the set of masked spatiotemporal patch indices, and $\mathbf{x}$ the visible input. The online encoder $E_\theta$ processes $\mathbf{x}$, and the predictor $P_\phi$ uses its outputs together with position-aware mask tokens $\boldsymbol{\Delta}_{\mathcal{M}}$ to predict the targets produced by the EMA encoder $E_{\bar{\theta}}$
\begin{equation}
\mathcal{L} =
\frac{1}{|\mathcal{M}|}
\sum_{i\in\mathcal{M}}
\left\|
P_\phi \left(
E_\theta(\mathbf{x}),
\boldsymbol{\Delta}_{\mathcal{M}}
\right)_i -
\operatorname{sg}\left(
E_{\bar{\theta}}(\mathbf{y})_i
\right)
\right\|_1.
\end{equation} 
Here, $\operatorname{sg}(\cdot)$ denotes stop-gradient, and $\bar{\theta}$ is updated by an EMA of $\theta$. V-JEPA provides dense spatiotemporal supervision, but its patch representations are not explicitly shaped as the state interface required for downstream world modeling and planning. We retain these dense targets while pretraining compact register states that are reused for future prediction and planning.

\subsection{Motion-Oriented Register Pretraining}
\label{sec:representation_pretraining}

Given a video clip $\mathbf{I}\in\mathbb{R}^{T\times H\times W\times 3}$, we tokenize it into non-overlapping tubelets of size $\tau\times p\times p$, yielding a token grid $\mathbf{P}\in\mathbb{R}^{N_t\times N_h\times N_w\times D}$, where $N_t=T/\tau$, $N_h=H/p$, and $N_w=W/p$. For each temporal chunk $k$, we denote its $N_p=N_hN_w$ patch tokens by $\mathbf{P}_k\in\mathbb{R}^{N_p\times D}$ and introduce $N_r\ll N_p$ learnable registers that interact with them through self-attention.

Patch tokens retain dense visual supervision during pretraining, while only the contextualized registers are reused as the state interface for future modeling and planning. We learn these motion-oriented registers through three complementary designs: ego-motion conditioning, register-mediated masked patch prediction, and blockwise register masking.

\PAR{Motion-Oriented Register States.}
We condition the registers on temporally aligned ego-motion. For adjacent frames, we define the relative ego-motion as
\begin{equation}
\mathbf{a}_t
=
\left(\Delta x_t,\Delta y_t,\Delta\psi_t\right)^\top,
\qquad
t=1,\ldots,T-1,
\end{equation}
where the motion from frame $t$ to $t+1$ is expressed in the ego coordinate system at frame $t$, and $\Delta x_t$, $\Delta y_t$, and $\Delta\psi_t$ denote the two translation components and the heading change. We align these frame-level motions with temporal chunks to obtain
$\widetilde{\mathbf{A}}
=
[\widetilde{\mathbf{a}}_k^\top]_{k=1}^{N_t}$.
Under the default setting $\tau=2$, $\widetilde{\mathbf{a}}_k=\mathbf{a}_{2k-1}$; for larger $\tau$, consecutive motions are composed in $\mathrm{SE}(2)$.

Let $\mathbf{E}^{\mathrm{reg}}\in\mathbb{R}^{N_t\times N_r\times D}$ denote all register embeddings, and let $\mathbf{W}_a\in\mathbb{R}^{3\times D}$ be a learnable action projection matrix. The motion-oriented input to the $r$-th register in the $k$-th chunk is defined as
\begin{equation}
\mathbf{R}^{0}_{k,r}
=
\mathbf{E}^{\mathrm{reg}}_{k,r}
+
\widetilde{\mathbf{a}}_k^\top\mathbf{W}_a,
\quad
k=1,\ldots,N_t,
\quad
r=1,\ldots,N_r,
\end{equation}
where $\mathbf{E}^{\mathrm{reg}}_{k,r}$ is the learnable embedding of the $r$-th register in the $k$-th chunk. Registers within each chunk share the same motion condition. Injecting aligned ego motion encourages the state space to capture factors related to scene dynamics and state evolution before downstream world modeling.

\paragraph{Two-Branch Pretraining with Shared Encoders.}
Building on the motion-oriented registers, we pretrain two complementary branches, namely Register-Mediated Masked Patch Prediction and Blockwise Register Masking. Both share the online encoder $E_\theta$, predictor $P_\phi$, and EMA target encoder $E_{\bar{\theta}}$. Patch prediction removes a subset of patch tokens while retaining all registers. Register masking retains all patches while removing a subset of registers. Their losses jointly optimize the shared model and constrain a common register state space. We describe each branch below.

\PAR{Register-Mediated Masked Patch Prediction.}
Simply introducing registers does not ensure that dense patch-level supervision effectively shapes them, since masked patch prediction can still rely directly on contextualized patch features. We therefore use the contextualized registers as the sole visual context for predicting masked patch representations.

Given the patch grid of size $N_t\times N_h\times N_w$, we sample contiguous spatiotemporal blocks with masked indices $\mathcal{M}_p$ and visible complement $\overline{\mathcal{M}}_p$. The online encoder processes the visible patches together with all action-conditioned registers
\begin{equation}
(\mathbf{P}^{p}_c,\mathbf{R}^{p}_c)
=
E_\theta\!\left(
\mathbf{P}_{\overline{\mathcal{M}}_p},
\mathbf{R}^{0}
\right).
\end{equation}
where $\mathbf{P}^{p}_c$ and $\mathbf{R}^{p}_c$ denote the contextualized patch and register outputs, respectively. Self-attention integrates visible patch information into $\mathbf{R}^{p}_c$. We then discard $\mathbf{P}^{p}_c$, and predict the masked patch representations using only $\mathbf{R}^{p}_c$ and position-aware queries
\begin{equation}
\widehat{\mathbf{H}}^{p}_{\mathcal{M}_p}
=
P_\phi\!\left(
\mathbf{Q}^{p}_{\mathcal{M}_p},
\mathbf{R}^{p}_c
\right),
\end{equation}
where $\mathbf{Q}^{p}_{\mathcal{M}_p}$ follows the same spatiotemporal indices as $\mathcal{M}_p$. 

The supervision targets $\mathbf{H}^{p}_{\mathcal{M}_p}$ are obtained from the EMA target encoder $E_{\bar{\theta}}$, which processes the complete patch sequence with the corresponding action-conditioned registers
\begin{equation}
(\mathbf{H}^{p},\mathbf{H}^{r})
=
E_{\bar{\theta}}\!\left(
\mathbf{P},
\mathbf{R}^{0}
\right),
\quad
\mathbf{H}^{p}_{\mathcal{M}_p}
=
\operatorname{sg}\!\left(
\mathbf{H}^{p}[\mathcal{M}_p]
\right),
\label{eq6}
\end{equation}
where $\mathbf{H}^{p}[\mathcal{M}_p]$ is the target patches at the masked indices. Each masked patch retains a spatially indexed latent target, while all visual context for prediction passes through the register bottleneck. This encourages the compact state to retain the information needed to recover patch latents.

\PAR{Blockwise Register Masking.}
The patch-prediction branch encourages registers to retain sufficient visual information but does not directly impose predictive structure on them. We therefore organize the registers as an $N_t\times N_r$ temporal chunk-slot grid and mask contiguous blocks $\mathcal{M}_r$, with $\overline{\mathcal{M}}_r$ denoting the visible complement. The online encoder processes all patches and visible registers
\begin{equation}
(\mathbf{P}^{r}_c,\mathbf{R}^{r}_c)
=
E_\theta\!\left(
\mathbf{P},
\mathbf{R}^{0}_{\overline{\mathcal{M}}_r}
\right).
\end{equation}
We discard contextualized patches $\mathbf{P}^{r}_c$ and use the shared predictor $P_\phi$ to recover the masked register states from $\mathbf{R}^{r}_c$
\begin{equation}
\widehat{\mathbf{H}}^{r}_{\mathcal{M}_r}
=
P_\phi\!\left(
\mathbf{Q}^{r}_{\mathcal{M}_r},
\mathbf{R}^{r}_c
\right),
\end{equation}
where $\mathbf{Q}^{r}_{\mathcal{M}_r}$ corresponds one-to-one with the masked registers. The targets are selected from the full register latents $\mathbf{H}^{r}$ produced by the EMA target encoder in Eq.~\ref{eq6}
\begin{equation}
\mathbf{H}^{r}_{\mathcal{M}_r}
=
\operatorname{sg}\!\left(
\mathbf{H}^{r}[\mathcal{M}_r]
\right).
\end{equation}
By predicting masked registers during pretraining, we establish the register space itself as a prediction target, which the downstream world model directly reuses for future-state prediction, reducing the mismatch between pretraining and world modeling.

\subsection{Joint World Modeling and Planning}
After representation pretraining, we initialize the online and EMA target encoders with the pretrained weights and retain only the registers as the shared state interface during joint training. Given $T_o$ observed frames from a $T$-frame video, we define $N_t^o=T_o/\tau$ and $N_t^f=(T-T_o)/\tau$ observed and future chunks, corresponding to $N_o=N_t^oN_r$ and $N_f=N_t^fN_r$ register tokens, respectively.

\PAR{Future Register World Modeling.}
The online encoder receives only the observed video and ego-motion to produce the observed registers
\begin{equation}
\mathbf{Z}_o
=
E_\theta\!\left(
\mathbf{I}_{1:T_o},
\mathbf{A}_{1:T_o-1}
\right)
\in
\mathbb{R}^{B\times N_o\times D}.
\end{equation}
During training, the EMA target encoder processes the full video and its temporally aligned ego-motion. Let $\mathcal{T}_f$ denote the register indices after the observation window. The complete target registers and the corresponding future targets are given by
\begin{equation}
\mathbf{H}
=
E_{\bar{\theta}}\!\left(
\mathbf{I}_{1:T},
\mathbf{A}_{1:T-1}
\right)
\in\mathbb{R}^{B\times(N_o+N_f)\times D},
\quad
\mathbf{H}_f
=
\operatorname{sg}\!\left(
\mathbf{H}[\mathcal{T}_f]
\right)
\in\mathbb{R}^{B\times N_f\times D}.
\end{equation}
The future video and ego-motion are used only to construct training targets and are not provided to the online branch or inference path.

We assign one learnable query per future temporal
chunk--register slot pair, yielding $\mathbf{Q}_f\in\mathbb{R}^{N_f\times D}$. To maintain consistency between training and inference, the motion condition $\widetilde{\mathbf{A}}^o$ retains the chunk-level ego-motion within the observation window and zero-pads all future positions. Given the current ego state $\mathbf{S}_o$, the world model $F_\psi$ jointly processes the observed registers and future queries
\begin{equation}
\widehat{\mathbf{Z}}_f
=
F_\psi\!\left(
\mathbf{Z}_o,
\mathbf{Q}_f,
\widetilde{\mathbf{A}}^o,
\mathbf{S}_o
\right)
\in
\mathbb{R}^{B\times N_f\times D}.
\end{equation}
The outputs at the future-query positions are ordered consistently with $\mathbf{H}_f$. All future queries interact with the observed registers and with one another, allowing the model to predict all future states in a single non-autoregressive pass.

\PAR{Trajectory Generation and Scoring.}
We concatenate the observed and predicted future registers into a shared scene memory $\mathbf{Z}\in\mathbb{R}^{B\times(N_o+N_f)\times D}$.
Conditioned on $\mathbf{Z}$ and the current ego state $\mathbf{S}_o$, the trajectory generator $D_\gamma$ maps $K$ learnable trajectory queries $\{\mathbf{q}_k^\tau\}_{k=1}^{K}$ to candidate trajectories
\begin{equation}
\widehat{\boldsymbol{\tau}}^{(k)}
=
D_\gamma\left(
\mathbf{q}_k^\tau,
\mathbf{Z},
\mathbf{S}_o
\right),
\quad
k=1,\ldots, K.
\end{equation}
An independent scorer $S_\omega$ evaluates the candidate trajectories using the same scene memory
\begin{equation}
\widehat{\mathbf{V}}
=
S_\omega\!\left(
\operatorname{sg}(\widehat{\boldsymbol{\tau}}),
\mathbf{Z}
\right)
\in
[0,1]^{B\times K\times C_s},
\end{equation}
where $\widehat{\mathbf{V}}$ contains $C_s$ quality scores per candidate, with targets constructed from the NAVSIM/PDM driving criteria~\citep{navsim}. Stop-gradient prevents the scoring loss from directly updating the trajectory generator through the predicted trajectories, while preserving gradients through $\mathbf{Z}$. Thus, the same register memory supports future prediction, trajectory generation, and scoring.

\subsection{Training Objectives and Inference}

\PAR{Stage 1.}
We supervise the masked patch and register predictions with $\ell_1$ losses
\begin{equation}
\mathcal{L}_{\mathrm{pretrain}}
=
\mathcal{L}_{\mathrm{patch}}
+
\mathcal{L}_{\mathrm{reg}}=
\frac{1}{|\mathcal{M}_p|}
\left\|
\widehat{\mathbf{H}}^p_{\mathcal{M}_p}
-
\mathbf{H}^p_{\mathcal{M}_p}
\right\|_1
+
\frac{1}{|\mathcal{M}_r|}
\left\|
\widehat{\mathbf{H}}^r_{\mathcal{M}_r}
-
\mathbf{H}^r_{\mathcal{M}_r}
\right\|_1.
\end{equation}
The online encoder and predictor are optimized by this objective, while the target encoder is updated by EMA as
$\bar{\theta}\leftarrow\mu\bar{\theta}+(1-\mu)\theta$, where $\mu\in[0,1)$ is the EMA momentum.

\PAR{Stage 2.}
Future register prediction is supervised by the EMA targets
\begin{equation}
\mathcal{L}_{\mathrm{pred}}
=
\frac{1}{N_f}
\left\|
\widehat{\mathbf{Z}}_f-\mathbf{H}_f
\right\|_1.
\end{equation}
The trajectory generator uses a minimum-over-$K$ loss
$\mathcal{L}_{\mathrm{traj}}$, while the scorer is trained with
$\mathcal{L}_{\mathrm{score}}$ to match oracle scores derived from the
NAVSIM/PDM driving criteria~\citep{navsim}. Their detailed definitions are provided in Appendix~\ref{loss_add}. The joint objective is
\begin{equation}
\mathcal{L}_{\mathrm{joint}}
=
\lambda_{\mathrm{pred}}\mathcal{L}_{\mathrm{pred}}
+
\lambda_{\mathrm{traj}}\mathcal{L}_{\mathrm{traj}}
+
\lambda_{\mathrm{score}}\mathcal{L}_{\mathrm{score}},
\end{equation}
where all loss weights are set to 1. The online encoder, world model, trajectory generator, and scorer are optimized jointly, while the target encoder is updated only by EMA. The oracle target scores are used only during training.

\PAR{Inference.}
At test time, the online encoder produces the observed registers, the world model predicts all future registers, and the planner generates and scores $K$ candidate trajectories from the resulting scene memory. We select the highest-scoring candidate for execution.

\section{Experiments}

\PAR{Datasets, Benchmarks \& Metrics.}
We pretrain on OpenScene~\citep{openscene}, a compact redistribution of nuPlan~\citep{nuplan} comprising approximately 120 hours of urban driving at 2~Hz, and jointly train the world model and planner on NAVSIM \texttt{navtrain}. We evaluate on the \texttt{navtest} splits of NAVSIM v1 and v2~\citep{navsim,navsim2}, reporting the Predictive Driver Model Score (PDMS) and Extended Predictive Driver Model Score (EPDMS) under their respective protocols. We further evaluate closed-loop driving in HUGSIM~\citep{hugsim}, where actions affect subsequent states and observations, reporting Route Completion (RC) and HUGSIM Driving Score (HD-Score). Metric details are provided in Appendices~\ref{navsim_evaluation} and \ref{hugsim_evaluation}.

\PAR{Implementation Details.} We follow Transfuser~\citep{transfuser} in using an input resolution of $256 \times 1024$, formed by stacking the front-left, front, and front-right images. The first-stage encoder is a ViT-L initialized from V-JEPA 2~\citep{V-jepa2} weights, with a temporal tubelet size of 2 frames and 8 registers per tubelet. We pretrain for 22,000 steps on OpenScene using AdamW on 32 NVIDIA A100 GPUs with a global batch size of 256, an initial learning rate of $1.5 \times 10^{-4}$ cosine-decayed to $10^{-6}$, and weight decay of $0.04$. The target-encoder EMA momentum increases from $0.996$ to $0.999$. 
For joint training, only the online and target encoders are initialized from the first stage, while the world model and planner are trained from scratch. The model observes 4 frames and predicts the subsequent 8 frames at $2$ Hz. We use $K=64$ trajectory queries. We train on \texttt{navtrain} for 10 epochs using 8 NVIDIA RTX 4090 GPUs with a global batch size of 128 and a base learning rate of $2 \times 10^{-4}$, applying a $0.1$ multiplier to the online encoder.

\subsection{Main Results}

\begin{table*}[t]
\centering
\captionsetup{skip=1.5pt}
\caption{
\textbf{Comparison on the NAVSIM-v2 benchmark.} EPDMS$^{*}$ uses the original NAVSIM-v2 evaluator, while EPDMS uses the official implementation with corrected human-behavior filtering.
}

\label{tab:navsim_v2_results}
\setlength{\tabcolsep}{4.5pt}
\renewcommand{\arraystretch}{0.95}
\setlength{\aboverulesep}{0.25ex}
\setlength{\belowrulesep}{0.25ex}

\resizebox{\textwidth}{!}{%
\begin{tabular}{@{}ll*{11}{c}@{}}
\toprule
Method
& Venue
& NC$\uparrow$
& DAC$\uparrow$
& DDC$\uparrow$
& TLC$\uparrow$
& EP$\uparrow$
& TTC$\uparrow$
& LK$\uparrow$
& HC$\uparrow$
& EC$\uparrow$
& EPDMS$^{*}\uparrow$
& EPDMS$\uparrow$ \\
\midrule

\multicolumn{13}{@{}l}{\textbf{End-to-End-based Methods}} \\

TransFuser~\citep{transfuser}
& TPAMI'22
& 96.9 & 89.9 & 97.8 & 99.7 & 87.1
& 95.4 & 92.7 & 98.3 & 87.2 & 76.7 & -- \\

ARTEMIS~\citep{artemis}
& RAL'25
& 98.3 & 95.1 & 98.6 & 99.8 & 87.6
& 97.4 & 96.5 & 98.3 & 89.1 & 83.1 & -- \\

Hydra-MDP++~\citep{hydra++}
& arXiv'25
& 98.8 & 97.8 & 99.1 & \textbf{100.0} & 84.0
& 95.3 & 70.1 & -- & \textbf{96.8} & 84.1 & -- \\

DiffusionDrive~\citep{liao2025diffusiondrive}
& CVPR'25
& 98.2 & 95.9 & 99.4 & 99.8 & 87.5
& 97.3 & 96.8 & 98.3 & 87.7 & -- & 84.5 \\

Drive-JEPA~\citep{drive-jepa}
& arXiv'26
& 98.4 & 98.6 & 99.1 & 99.8 & 88.4
& 97.8 & 97.6 & 97.9 & 84.8 & 87.8 & -- \\

SparseDriveV2~\citep{sun2026sparsedrivev2}
& ECCV'26
& 98.1 & 98.1 & 99.6 & 99.8 & 91.1
& 97.3 & 96.9 & 98.2 & 78.4 & 86.7 & 90.1 \\

DrivoR~\citep{drivor}	
& CVPR’26	
&99.3 &\textbf{99.1}	& 99.7	& 99.7	&87.6	&\textbf{98.9}	
&95.8	&97.9	&73.4	& --	&90.3	\\		

NTR~\citep{ntr}
& arXiv'26
& \textbf{99.4} & 99.0 & 99.7 & 99.7 & 88.3
& \textbf{98.9} & 95.5 & 98.0 & 76.9 & -- & 90.9 \\

\midrule
\multicolumn{13}{@{}l}{\textbf{VLA-based Methods}} \\

ReCogDrive~\citep{recogdrive}
& ICLR'26
& 98.3 & 95.2 & 99.5 & 99.8 & 87.1
& 97.5 & 96.6 & 98.3 & 86.5 & 83.6 & -- \\

WAM-Flow~\citep{wam-flow}
& CVPR'26
& 98.5 & 94.5 & 99.5 & 99.8 & 86.9
& 96.8 & 97.4 & 97.6 & 73.9 & 84.7 & -- \\

DriveVLA-W0~\citep{li2026drivevla}
& ICLR'26
& 98.5 & \textbf{99.1} & 98.0 & 99.7 & 86.4
& 98.1 & 93.2 & 97.9 & 58.9 & 86.1 & -- \\

DriveWorld-VLA~\citep{driveworld-vla}
& ICML'26
& 98.6 & \textbf{99.1} & 99.6 & 99.8 & 87.4
& 97.9 & 97.0 & 97.8 & 78.6 & -- & 86.8 \\

CoWorld-VLA~\citep{coworld-vla}
& arXiv'26
& 99.1 & 97.0 & 99.6 & 99.9 & 87.9
& 98.5 & \textbf{97.7} & 98.2 & 86.2 & 86.2 & 90.0 \\

\midrule
\multicolumn{13}{@{}l}{\textbf{World-Model-based Methods}} \\

WoTE~\citep{WoTE}
& ICCV'25 
&98.5 &96.8 &98.8 &99.8 &86.1 &97.9 
&95.5 &98.3 &82.9 &-- &87.7\\

DreamerAD~\citep{dreamerad}
& ECCV'26
& 98.0 & 97.2 & 99.5 & 99.8 & 87.8
& 97.4 & 97.5 & 98.3 & 72.4 & -- & 87.7 \\

DriveLaW~\citep{drivelaw} 
&CVPR'26 &98.7 &96.9 &99.6 &99.8 &87.5 
&98.3 &97.6 &\textbf{98.4} &77.4 &-- &88.6\\

EponaV2~\citep{eponav2}
& arXiv'26
&98.5 &97.4 &99.5 &99.9 &87.9 &98.1 &\textbf{97.7} &98.2 &77.4 & -- &88.9 \\

Latent-WAM~\citep{latent-wam}
& arXiv'26
& 98.1 & 97.3 & 99.6 & 99.8 & 87.7
& 97.3 & 97.6 & 98.1 & 87.3 & -- & 89.3 \\

Discrete-WAM~\citep{discrete-wam}
& arXiv'26
& 98.5 & 98.2 & 99.7 & 99.8 & 90.5
& 97.9 & 97.2 & 98.3 & 78.1 & 87.0 & 90.4 \\

\textbf{Ours}
& --
& 98.9 & 99.0 & \textbf{99.8} & 99.6 & \textbf{91.7}
& 98.6 & 97.1 & \textbf{98.4} & 84.8 & -- & \textbf{92.5} \\

\bottomrule
\end{tabular}%
}
\vspace{-8pt}
\end{table*}

\PAR{Results on NAVSIM-v2.}
Tab.~\ref{tab:navsim_v2_results} compares MORE with SOTA end-to-end, VLA-based, and world-model-based approaches. MORE achieves the highest EPDMS of 92.5, exceeding DrivoR~\citep{drivor}, CoWorld-VLA~\citep{coworld-vla}, and Discrete-WAM~\citep{discrete-wam} by 2.2, 2.5, and 2.1, respectively. It also achieves the best Ego Progress (EP), Driving-Direction Compliance (DDC), and History Comfort (HC), indicating strong driving progress without sacrificing directional compliance or comfort. These results show that compact register states learned during video pretraining can provide an effective shared interface for future modeling and planning.

\begin{table*}[t]
    \centering
    \captionsetup{skip=1.5pt}
    \caption{\textbf{Photorealistic closed-loop evaluation on HUGSIM.} All results use the corrected evaluator with fixes to trajectory-derived heading computation and related issues.}
    
    \label{tab:hugsim}
    \setlength{\tabcolsep}{5.5pt}
    \renewcommand{\arraystretch}{0.92}
    \setlength{\aboverulesep}{0.25ex}
    \setlength{\belowrulesep}{0.25ex}
    \resizebox{\textwidth}{!}{
    \begin{tabular}{lcccccccccc}
        \toprule
        \multirow{2}{*}{Method}
        & \multicolumn{5}{c}{RC$\uparrow$}
        & \multicolumn{5}{c}{HD-Score$\uparrow$} \\
        \cmidrule(lr){2-6}
        \cmidrule(lr){7-11}
        & E & M & H & X & Avg.
        & E & M & H & X & Avg. \\
        \midrule
        UniAD~\citep{hu2023planning}
        & 58.6 & 41.2 & 40.4 & 26.0 & 40.6
        & 48.7 & 29.5 & 27.3 & 14.3 & 28.9 \\

        VAD~\citep{vad}
        & 38.7 & 27.0 & 25.5 & 23.0 & 27.9
        & 24.3 & 9.9 & 10.4 & 8.2 & 12.3 \\

        LTF~\citep{transfuser}
        & 68.4 & 40.7 & 36.9 & 25.5 & 41.4
        & 52.8 & 24.6 & 19.8 & 8.1 & 24.8 \\

        GTRS-Dense~\citep{GTRS}
        & 64.2 & 50.0 & 20.7 & 22.3 & 38.0
        & 55.5 & 39.0 & 11.7 & 14.3 & 28.6 \\

        DrivoR~\citep{drivor}
        &85.9  &36.6  &34.1  &30.4 &43.7
        &77.2 & 19.3 &16.8 &13.9 &28.1 \\
        
        Latent-WAM~\citep{latent-wam}
        & 84.2 & 42.5 & 30.6 & 35.5 & 45.9
        & 72.5 & 24.0 & 12.2 & 18.1 & 28.9 \\

        \textbf{Ours}
        &88.2  &59.9  &35.7  &25.6 &\textbf{51.6}
        &77.4 &49.4  &21.7  &11.1  &\textbf{39.4}\\
        \bottomrule
    \end{tabular}
    }

    \vspace{2pt}
    \parbox{\textwidth}{
        \raggedleft
        \footnotesize
        \textit{E: Easy, M: Medium, H: Hard, X: Extreme.}
    }
\vspace{-14pt}
\end{table*}

\PAR{Results on HUGSIM.}
We further conduct zero-shot closed-loop evaluation on 436 HUGSIM scenarios spanning four difficulty levels, without benchmark-specific training or fine-tuning. As shown in Tab.~\ref{tab:hugsim}, MORE achieves the highest average Route Completion (RC) of 51.6 and HD-Score of 39.4. These results show that the compact register states generalize beyond NAVSIM and provide effective predictive context for long-horizon driving decisions.

\PAR{Results on NAVSIM-v1.} As shown in Tab.~\ref{tab:main_results}, MORE achieves 93.2 PDMS On NAVSIM-v1 \texttt{navtest}, outperforming the strong baselines iPad~\citep{ipad}, SGDrive~\citep{sgdrive}, and DriveFuture~\citep{drivefuture} by 1.5, 2.1, and 2.5, respectively. It also achieves the highest Ego Progress (EP) of 90.7 while remaining competitive in safety and compliance. These results show that compact register states provide effective planning representations across benchmarks.

\begin{table*}[t]
\centering
\captionsetup{skip=1.5pt}
\caption{\textbf{Comparison on the NAVSIM-v1 benchmark.} C and L denote camera and LiDAR input, respectively. $^\dagger$: The results are dependent on perception-based annotation.}
\label{tab:main_results}

\begingroup 
\setlength{\tabcolsep}{5pt}
\renewcommand{\arraystretch}{0.9} 

\setlength{\aboverulesep}{0.25ex} 
\setlength{\belowrulesep}{0.25ex} 

\resizebox{\textwidth}{!}{
\begin{tabular}{@{}llccccccc@{}}
\toprule
Method & Venue & Sensors
& NC$\uparrow$
& DAC$\uparrow$
& TTC$\uparrow$
& C$\uparrow$
& EP$\uparrow$
& PDMS$\uparrow$ \\
\midrule

\multicolumn{9}{@{}l}{\textbf{End-to-End-based Methods}} \\
UniAD$^\dagger$~\citep{hu2023planning}
& CVPR'23
& 6C 
& 97.8 & 91.9 & 92.9 & 100.0 & 78.8 & 83.4 \\

Transfuser~\citep{transfuser}
& TPAMI'22
& 3C\&L
& 97.7 & 92.8 & 92.8 & 100.0 & 79.2 & 84.0 \\

Hydra-MDP++~\citep{hydra++}
& arXiv'25
& 3C
& 98.6 & 98.6 & 95.1 & 100.0 & 85.7 & 91.0 \\

DiffusionDrive$^\dagger$~\citep{liao2025diffusiondrive}
& CVPR'25
& 3C\&L
& 98.2 & 96.2 & 94.7 & 100.0 & 82.2 & 88.1 \\

Drive-JEPA~\citep{drive-jepa}
&arXiv'26
& 1C
&98.7 &96.2 &95.5 &100 &82.9 &89.0 \\

iPad$^\dagger$~\citep{ipad}
& RAL’26
& 4C
& 98.6 & 98.3 & 94.9 & 100.0 & 88.0 & 91.7 \\

\midrule
\multicolumn{9}{@{}l}{\textbf{VLA-based Methods}} \\
AutoVLA~\citep{autovla}
& NeurIPS'25
& 3C
& 98.4 & 95.6 & 98.0 & 99.9 & 81.9 & 89.1 \\

UniVLA~\citep{UniVLA}
& ICLR'26
& 1C
& 98.3 & 93.8 & 94.2 & 100.0 & 80.0 & 85.6 \\

DriveVLA-W0~\citep{li2026drivevla}
& ICLR'26
& 1C
& 98.7 & 99.1 & 95.3 & 99.3 & 83.3 & 90.2 \\

ReCogDrive~\citep{recogdrive}
& ICLR'26
& 3C
& 97.9 & 97.3 & 94.9 & 100.0 & 87.3 & 90.8 \\

ExploreVLA~\citep{explorevla}
& ECCV'26
& 1C
& 98.8 & 98.4 & 96.5 & 99.9 & 83.5 & 90.4 \\

SGDrive$^\dagger$~\citep{sgdrive}
& CVPR'26
& 1C
& 98.6 & 97.8 & 96.2 & 100.0 & 85.8 & 91.1 \\

\midrule
\multicolumn{9}{@{}l}{\textbf{World-Model-based Methods}} \\

LAW~\citep{li2025enhancing}
& ICLR'25
& 1C
& 96.4 & 95.4 & 88.7 & 99.9 & 81.7 & 84.6 \\

World4Drive~\citep{zheng2025world4drive}
& ICCV'25
& 3C
& 97.4 & 94.3 & 92.8 & 100.0 & 79.9 & 85.1 \\

WoTE$^\dagger$~\citep{WoTE}
& ICCV'25
& 3C\&L 
& 98.5 & 96.8 & 94.9 & 99.9 & 81.9 & 88.3 \\

WorldRFT~\citep{worldrft}
& AAAI'26
& 3C
& 97.8 & 96.8 & 94.0 & 100.0 & 81.7 & 87.8 \\

PWM~\citep{PWM}
&NeurIPS’26
&1C	
&98.6 	&95.9 	&95.4 	&100.0 	&81.8 	&88.1\\

ResWorld$^\dagger$~\citep{resworld}
& ICLR'26
& 3C\&L 
& 98.9 & 96.5 & 95.6 & 100.0 & 83.1 & 89.0 \\

DriveLaW~\citep{drivelaw}
& CVPR'26	
& 1C	
& 99.0 	&97.1 	&96.7 	&100.0 	&81.3 	&89.1\\ 

EponaV2~\citep{eponav2}
& arXiv'26
& 1C
& 98.6 & 97.9 & 95.7 & 100.0 & 84.8 & 90.4 \\

DriveFuture~\citep{drivefuture} 
& arXiv'26
& 6C
&98.8 &99.1 &95.4 &100.0 &84.2 &90.7 \\

\textbf{Ours}
& --
& 3C
& 98.8 & 98.4 & 95.6 & 100.0
& 90.7
& \textbf{93.2} \\
\bottomrule
\end{tabular}
}

\endgroup 
\vspace{-7pt}
\end{table*}

\subsection{Ablation Study}

\begin{table}[t]
\centering
\captionsetup{skip=1.5pt}
\caption{\textbf{Ablation Study on NAVSIM \texttt{navtest}.} All variants use only the front-camera input.}
\label{tab:ablation1}
\setlength{\tabcolsep}{3.8pt}
\renewcommand{\arraystretch}{1}

\resizebox{\linewidth}{!}{
\begin{tabular}{clccccccccc}
\toprule
\multirow{2}{*}{Case}
& \multirow{2}{*}{Configuration}
& \multicolumn{4}{c}{Representation Pre-training}
& \multicolumn{2}{c}{Joint Training}
& \multicolumn{2}{c}{Performance} \\
\cmidrule(lr){3-6}
\cmidrule(lr){7-8}
\cmidrule(lr){9-10}
&
& +Reg.
& Motion
& Patch Ctx.
& Reg. Mask
& State
& Plan. Mem.
& EPDMS
& PDMS \\
\midrule

\#1
& V-JEPA Baseline
& \xmark
& \xmark
& Patch
& \xmark
& Patch
& Obs.+Pred.
& 84.9
& 85.9 \\

\#2
& V-JEPA + Patch Motion
& \xmark
& Patch
& Patch
& \xmark
& Patch
& Obs.+Pred.
& 89.4
& 89.2 \\

\#3
& Joint-Initialized Reg.
& \xmark
& \xmark
& Patch
& \xmark
& Reg.$^\ast$
& Obs.+Pred.
& 85.7
& 86.0 \\

\#4
& Reg. Interaction Only
& \cmark
& \xmark
& Patch
& \xmark
& Reg.
& Obs.+Pred.
& 85.6
& 86.3 \\

\#5
& Register-Mediated Pred.
& \cmark
& \xmark
& Reg.
& \xmark
& Reg.
& Obs.+Pred.
& 88.5
& 89.3 \\

\#6
& Full w/o Motion
& \cmark
& \xmark
& Reg.
& \cmark
& Reg.
& Obs.+Pred.
& 89.1
& 90.2 \\

\#7
& Full w/o Future Pred.
& \cmark
& Reg.
& Reg.
& \cmark
& Reg.
& Obs.
& 91.1
& 92.4 \\

\midrule

\#8
& \textbf{Full (Ours)}
& \cmark
& Reg.
& Reg.
& \cmark
& Reg.
& Obs.+Pred.
& \textbf{91.8}
& \textbf{93.0} \\

\bottomrule
\end{tabular}
}

\vspace{3pt}
\begin{minipage}{\linewidth}
\scriptsize
\raggedright
``+Reg.'' indicates whether registers are used during pre-training;
``Motion'' denotes the token type receiving ego-motion conditioning during representation pre-training.
``Patch Ctx.'' denotes the context used for masked patch prediction;
``Reg. Mask'' denotes the blockwise register masking;
and ``Plan. Mem.'' denotes the states provided to the planner.
$^\ast$ Registers are introduced and initialized only during joint training.
\end{minipage}
\vspace{-10pt}
\end{table}

\PAR{Learning the Register State Interface.} Tab.~\ref{tab:ablation1} shows that simply introducing registers is insufficient. Initializing them only during joint training (\#3) or allowing them to interact with patches during pretraining while retaining patch-based prediction (\#4) brings limited gains over the baseline (\#1). Using registers as the sole visual context for masked patch prediction (\#5 VS. \#4) raises EPDMS from 85.6 to 88.5 and PDMS from 86.3 to 89.3. Blockwise register masking (\#6) further improves them to 89.1 and 90.2. These results show that two branches provide complementary supervision for learning information-rich and predictable compact states.

\PAR{Action Conditioning.}
Injecting ego motion improves both the patch-based baseline (\#2 VS. \#1) and the register-based model (\#8 VS. \#6), confirming the effectiveness of motion-conditioned pretraining. Our model benefits more and clearly outperforms its patch-based counterpart (\#2), reaching 91.8 EPDMS and 93.0 PDMS. This suggests that the register bottleneck better preserves motion- and planning-relevant information while suppressing redundant visual details.

\PAR{Future-State Prediction.}
Finally, \#7 removes future-register prediction and supplies the planner with observed registers only, while the full model (\#8) uses predicted future registers. The resulting improvement in EPDMS and PDMS confirms that the predicted states capture decision-relevant scene evolution beyond the current observation and provide useful context for trajectory planning.

\begin{wraptable}{r}{0.45\columnwidth}
\vspace{-12pt}
\centering
\captionsetup{skip=1pt}
\caption{\textbf{Register-count ablation.}}
\label{tab:register_num}

\setlength{\tabcolsep}{2pt}
\renewcommand{\arraystretch}{0.9}
\resizebox{0.9\linewidth}{!}{
\begin{tabular}{@{}lcccc@{}}
\toprule
Register Num. & 4 & 8 & 16 & 32 \\
\midrule
EPDMS$\uparrow$ & 91.4 & \textbf{91.8} & 91.6 & 91.5 \\
\bottomrule
\end{tabular}
}
\vspace{-6pt}
\end{wraptable}

\PAR{Number of Registers.}
Tab.~\ref{tab:register_num} studies the number of registers allocated to each temporal chunk. Using 8 registers achieves the best EPDMS of 91.8, while increasing $N_r$ to 16 or 32 provides no further improvement. We use $N_r=8$ throughout our experiments.

\subsection{Visualization}

\begin{figure}[h]
\begin{center}
\includegraphics[width=\linewidth]{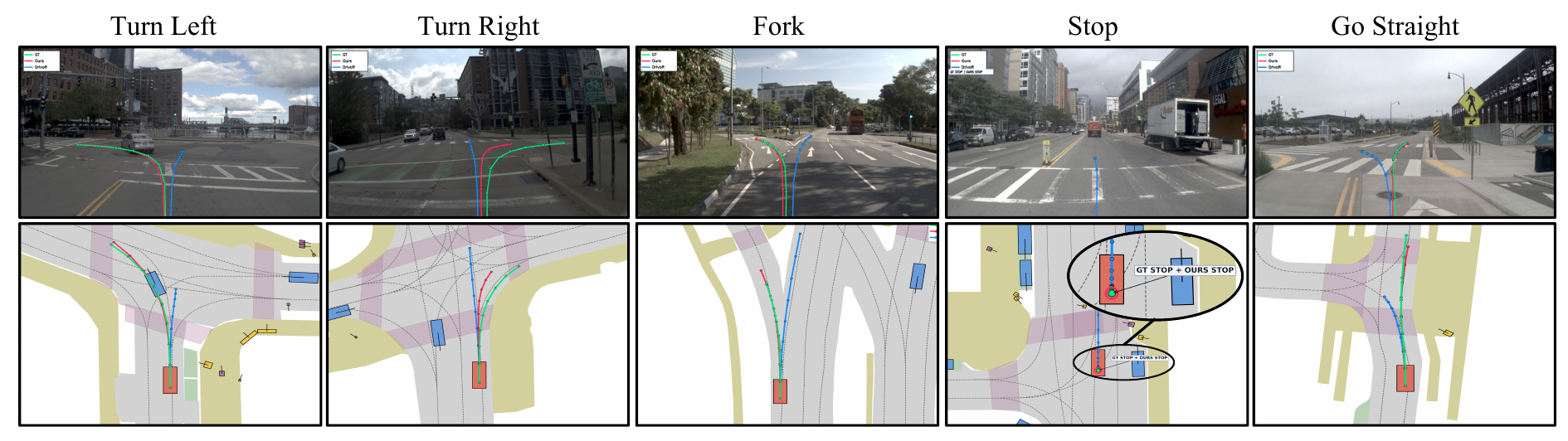}
\end{center}
\captionsetup{skip=1pt}
\caption{\textbf{Qualitative trajectory comparison.} 
Trajectories:
  \protect\tikz[baseline=-0.5ex]{\protect\shade[left color=green!70!black, right color=green!50!yellow] (0,-0.05) rectangle (0.4,0.05); \protect\fill[green!70!black] (0.2,0) circle (2.5pt);} GT,
  \protect\tikz[baseline=-0.5ex]{\protect\shade[left color=red!85!black, right color=red!65!white](0,-0.05) rectangle (0.4,0.05);\protect\fill[red] (0.2,0) circle (2.5pt);} MORE (Ours),
  \protect\tikz[baseline=-0.5ex]{\protect\shade[left color=cyan!80!blue, right color=cyan!50!blue] (0,-0.05) rectangle (0.4,0.05); \protect\fill[cyan!80!blue] (0.2,0) circle (2.5pt);} DrivoR.
}
\label{vis1}
\vspace{-2pt}
\end{figure}

Fig.~\ref{vis1} compares MORE and DrivoR across left and right turns, a fork, stopping, and straight driving. MORE closely follows the ground-truth trajectory across all scenarios, selecting the correct branches, stopping near the reference position, and maintaining route-aligned motion, whereas DrivoR exhibits larger deviations and fails to stop in the stopping scenario. These examples suggest that motion-oriented registers preserve decision-relevant scene dynamics, providing a compact state interface well aligned with downstream planning.

\section{Conclusion}

We present MORE, which learns motion-oriented registers during video pretraining as a compact state interface for driving world models and planning. During pretraining, ego-motion conditioning, register-mediated masked patch prediction, and blockwise register masking retain dense visual supervision while shaping informative, motion-sensitive, and predictive states. During joint training, the same register space supports non-autoregressive world modeling, trajectory generation, and scoring. Results on NAVSIM-v1, NAVSIM-v2, and HUGSIM demonstrate strong planning performance, low inference latency, and zero-shot closed-loop generalization. Overall, MORE shows that video pretraining can directly shape the compact predictive state required by world modeling and planning rather than leaving this adaptation to downstream training.

\bibliography{main}
\bibliographystyle{iclr2027_conference}

\appendix

\newpage
\section*{Appendix}

\startcontents[sections]
\printcontents[sections]{l}{1}{\setcounter{tocdepth}{2}}

\section{Overall}
\label{app:overall}

In this appendix, we provide supplementary materials supporting the main paper. Appendix~\ref{B} describes the datasets, trajectory and scoring losses, and further implementation details. Appendices~\ref{navsim_evaluation} and~\ref{hugsim_evaluation} detail the NAVSIM and HUGSIM evaluation protocols, respectively. Appendix~\ref{efficiency} presents the efficiency analysis, while Appendix~\ref{F} provides additional trajectory comparisons, register-similarity analysis, and representative failure cases.

\section{Datasets and More Implementation Details}
\label{B}
\subsection{Datasets}

\PAR{OpenScene.}
OpenScene is a redistribution of nuPlan driving logs sampled at 2 Hz, comprising approximately 120 hours of driving. We use its \texttt{trainval} logs for motion-oriented register pretraining. Consecutive windows start 1 frame (0.5 s) apart. Each window provides a 12-frame video clip to the model. No OpenScene \texttt{test} logs are used for pretraining.

\PAR{NAVSIM Training and Evaluation Splits.}
For joint training of the world model and planner, we load NAVSIM's \texttt{navtrain}. Frames are sampled at 2 Hz, and consecutive candidate windows start 1 frame (0.5 s) apart. After scene filtering and data-validity checks, the training set contains 103,288 samples. For each sample, the model observes 4 frames and uses the following 8 frames for future-state and trajectory supervision.

For evaluation, the NAVSIM's \texttt{navtest} filter selects 12,146 scenarios. We report PDMS under the NAVSIM-v1 protocol and EPDMS under the NAVSIM-v2 protocol on this same \texttt{navtest} set.

\PAR{HUGSIM.} HUGSIM is a photorealistic closed-loop simulator in which the ego vehicle's actions affect subsequent states and rendered observations. Its benchmark draws on reconstructed scenes from KITTI-360~\citep{kitti}, Waymo Open Dataset~\citep{waymo}, nuScenes~\citep{nuscenes}, and PandaSet~\citep{pandaset}. We evaluate the NAVSIM-trained model without HUGSIM-specific training or fine-tuning on 436 scenarios: 80 Easy, 157 Medium, 96 Hard, and 103 Extreme. This evaluation tests the policy under interactive, long-horizon driving rather than the non-reactive NAVSIM scoring protocol.

\subsection{Trajectory and Scoring Losses}
\label{loss_add}
Each of the $K$ trajectory queries produces a candidate
\begin{equation}
\widehat{\boldsymbol{\tau}}^{(k)}
=
\left\{
\left(
\widehat{\mathbf{p}}^{(k)}_m,
\widehat{\psi}^{(k)}_m
\right)
\right\}_{m=1}^{M},
\qquad
\boldsymbol{\tau}^{*}
=
\left\{
\left(
\mathbf{p}^{*}_m,
\psi^{*}_m
\right)
\right\}_{m=1}^{M},
\end{equation}
where $M$ is the prediction horizon, $\widehat{\mathbf{p}}^{(k)}_m,\mathbf{p}^{*}_m\in\mathbb{R}^{2}$ are the predicted and expert planar positions, and $\widehat{\psi}^{(k)}_m,\psi^{*}_m$ are their headings at step $m$. We define
$\mathbf{o}(\psi)=(\cos\psi,\sin\psi)^\top$
to represent heading without angular discontinuities. The minimum-over-$K$ trajectory loss is
\begin{equation}
\mathcal{L}_{\mathrm{traj}}
=
\min_{1\leq k\leq K}
\frac{1}{M}
\sum_{m=1}^{M}
\Bigg[\left\|
\widehat{\mathbf{p}}^{(k)}_m-\mathbf{p}^{*}_m
\right\|_1 +
\lambda_{\mathrm{yaw}}
\left\|
\mathbf{o}\!\left(\widehat{\psi}^{(k)}_m\right)
-
\mathbf{o}\!\left(\psi^{*}_m\right)
\right\|_1
\Bigg],
\end{equation}
where $\lambda_{\mathrm{yaw}}$ weights the heading term and is set to $1$.

For each candidate, the learned scorer predicts
$\widehat{\mathbf{V}}^{(k)}
=(\widehat{V}^{(k)}_c)_{c=1}^{C_s}$,
while the NAVSIM/PDM oracle evaluator~\citep{navsim} constructs the corresponding target
$\mathbf{V}^{*(k)}
=(V^{*(k)}_c)_{c=1}^{C_s}\in[0,1]^{C_s}$.
Here, $C_s$ is the number of supervised driving criteria. We optimize the scorer using
\begin{equation}
\mathcal{L}_{\mathrm{score}}
=
\frac{1}{K C_s}
\sum_{k=1}^{K}
\sum_{c=1}^{C_s}
\operatorname{BCE}\!\left(
\widehat{V}^{(k)}_c,
V^{*(k)}_c
\right),
\end{equation}
where
$\operatorname{BCE}(\widehat{v},v)
=
-v\log\widehat{v}
-(1-v)\log(1-\widehat{v})$.
Together with the world model loss $\mathcal{L}_{\mathrm{pred}}$, the stage 2 objective is
\begin{equation}
\mathcal{L}_{\mathrm{joint}}
=
\lambda_{\mathrm{pred}}\mathcal{L}_{\mathrm{pred}}
+
\lambda_{\mathrm{traj}}\mathcal{L}_{\mathrm{traj}}
+
\lambda_{\mathrm{score}}\mathcal{L}_{\mathrm{score}},
\end{equation}
where
$\lambda_{\mathrm{pred}}
=\lambda_{\mathrm{traj}}
=\lambda_{\mathrm{score}}=1$.
The EMA target encoder receives no gradients, and the oracle evaluator is used only to construct training targets.

\subsection{More Implementation Details}
\label{app:implementation_details}

\PAR{Input Processing.}
For each timestamp, we use synchronized front-left, front, and front-right images. Following TransFuser~\citep{transfuser}, we remove 28 pixels from the top and bottom of each image and crop 416 pixels from both horizontal sides of the two lateral views. The processed views are concatenated in left--front--right order, resized once to $256\times1024$. We apply no additional image augmentation to preserve the geometry of the stitched view.

\PAR{Stage1: Motion-Oriented Register Pretraining.}
We use a ViT-L video encoder with 24 Transformer blocks, a token dimension of 1,024, and 16 attention heads. Using a temporal tubelet size of 2 frames and a spatial patch size of $16\times16$, each 12-frame clip yields 6 temporal chunks and 48 registers. A three-dimensional ego-motion vector is linearly projected to the encoder dimension and added to the registers of its corresponding chunk before self-attention. The register-mediated masked patch prediction and blockwise register masking branches share a 12-layer Transformer predictor with a hidden dimension of 384 and 12 attention heads.

\PAR{Stage2: Joint World Modeling and Planning.}
Four observed frames yield two temporal chunks and 16 observed registers, while the eight future frames correspond to four chunks and 32 learned future queries. The trajectory generator and scorer each contain 4 Transformer layers with a hidden dimension of 256. We configure AdamW for 10 epochs, with the base learning rate $2\times10^{-4}$, followed by cosine decay to $10^{-6}$. The encoder parameter groups use a learning-rate multiplier of 0.1, giving scheduled rates of $2\times10^{-5}$ at the start and $10^{-7}$ at the end. Training uses bfloat16 precision.

\section{NAVSIM Evaluation}
\label{navsim_evaluation}

NAVSIM~\citep{navsim} evaluates predicted ego trajectories through a lightweight simulation-based protocol. The official simulator executes the trajectory using its vehicle model and evaluates it against map, route, and traffic-agent information.

\PAR{NAVSIM v1.}
NAVSIM v1 uses non-reactive simulation, in which the ego vehicle executes the predicted trajectory, while surrounding agents follow their recorded future trajectories. The Predictive Driver Model Score (PDMS) comprises five normalized subscores. No At-Fault Collisions (NC) penalizes collisions for which the ego vehicle is considered responsible, with a softer penalty for collisions with static objects. Drivable Area Compliance (DAC) checks whether the ego footprint remains within the drivable area. Ego Progress (EP) measures progress along the route center relative to a safe upper bound estimated by PDM-Closed. Time-to-Collision (TTC) checks whether the projected TTC under constant velocity and heading remains above a safety threshold. Comfort (C) verifies that trajectory acceleration and jerk remain within predefined bounds. The PDMS is
\begin{equation}
\mathrm{PDMS}
=
\mathrm{NC}
\times
\mathrm{DAC}
\times
\frac{
5\,\mathrm{EP}
+
5 \,\mathrm{TTC}
+
2 \,\mathrm{C}
}{12}.
\label{eq:pdms}
\end{equation}
NC and DAC act as multiplicative safety penalties, while EP, TTC, and C measure route progress, anticipated collision risk, and dynamic feasibility, respectively.

\PAR{NAVSIM v2.}
We apply the NAVSIM v2 metric~\citep{navsim2} to report the Extended PDMS (EPDMS). In addition to the v1 criteria, EPDMS evaluates Driving Direction Compliance (DDC), Traffic Light Compliance (TLC), Lane Keeping (LK), History Comfort (HC), and Extended Comfort (EC). Let $s_m^{\mathrm{A}}$ and $s_m^{\mathrm{H}}$ denote the agent and human-reference scores, respectively, for metric $m$. The official human-reference filter is defined as
\begin{equation}
\phi_m
=
\begin{cases}
1, & \text{if } s_m^{\mathrm{H}}=0,\\
s_m^{\mathrm{A}}, & \text{otherwise}.
\end{cases}
\label{eq:human_filter}
\end{equation}
This filter suppresses a metric-specific penalty when the corresponding human trajectory also violates that criterion. The EPDMS is computed as
\begin{equation}
\begin{aligned}
\mathrm{EPDMS}
={}&
\phi_{\mathrm{NC}}
\times
\phi_{\mathrm{DAC}}
\times
\phi_{\mathrm{DDC}}
\times
\phi_{\mathrm{TLC}}
\\
&\times
\frac{
5 \, \phi_{\mathrm{EP}}
+
5 \,\phi_{\mathrm{TTC}}
+
2 \,\phi_{\mathrm{LK}}
+
2 \,\phi_{\mathrm{HC}}
+
2 \,\phi_{\mathrm{EC}}
}{16}.
\end{aligned}
\label{eq:epdms}
\end{equation}
LK measures sustained deviation from the lane center, HC evaluates consistency with the observed motion history, and EC measures dynamic consistency across consecutive predictions. DDC and TLC penalize wrong-way driving and traffic-light violations.

\section{HUGSIM Closed-Loop Evaluation}
\label{hugsim_evaluation}

HUGSIM~\citep{hugsim} is a photorealistic, interactive closed-loop simulator constructed using 3D Gaussian Splatting~\citep{3DGS}. Unlike NAVSIM's per-scene trajectory evaluation, HUGSIM repeatedly queries the policy throughout an episode. The predicted trajectory is executed, the ego vehicle and surrounding actors are updated, and new observations are rendered from the resulting state. Prediction errors therefore affect subsequent inputs and may accumulate over time.

HUGSIM evaluates NC (No Collision), DAC, TTC, and C under its official protocol, while our main comparison reports Route Completion (RC) and HUGSIM Driving Score (HD-Score) for a consistent benchmark-level comparison, as they summarize task completion and overall closed-loop driving quality. RC measures the fraction of the predefined route completed by the ego vehicle. The official implementation computes RC from the furthest reached route-waypoint index, normalized by $0.9$ times the total waypoint count and clipped to $1$. At evaluated keyframe $t$, the driving-quality score is
\begin{equation}
\text{HD-Score}_t
=
\mathrm{NC}_t
\times
\mathrm{DAC}_t
\times
\frac{
5\,\mathrm{TTC}_t
+
2\,\mathrm{C}_t
}{7}.
\label{eq:hugsim_step_score}
\end{equation}
For an episode containing $N_{\mathrm{eval}}$ evaluated keyframes, the final score is
\begin{equation}
\text{HD-Score}
=
\mathrm{RC}
\times
\frac{1}{N_{\mathrm{eval}}}
\sum_{t=1}^{N_{\mathrm{eval}}}
\text{HD-Score}_t.
\label{eq:hugsim_hdscore}
\end{equation}
HD-Score therefore scales the average closed-loop driving quality by route completion, preventing a stationary but collision-free policy from receiving a high score.

\section{Efficiency analysis}
\label{efficiency}

\begin{table*}[t]
\centering
\captionsetup{skip=1.5pt}
\caption{\textbf{Efficiency.} We compare model parameters, peak GPU memory, and inference latency with respect to NAVSIM-v2 EPDMS.}
\label{tab:efficiency}
\footnotesize
\setlength{\tabcolsep}{4pt}
\renewcommand{\arraystretch}{1.2}
\begin{tabular}{@{}lccccccc@{}}
\toprule
Method
& Img. Size
& Obs. Frames
& Cams
& Params. $\downarrow$
& \shortstack{Peak Mem. $\downarrow$}
& \shortstack{Latency $\downarrow$}
& EPDMS$\uparrow$ \\
\midrule
Epona
& $512\!\times\!1024$ & 10 & 1
& 2,915.4 M& 14.70 GB & 900.0 ms & 85.1 \\

World4Drive
& $384\!\times\!640$ & 4 & 6
& 70.2 M& 1.98 GB & 414.5 ms & 84.8 \\

DriveVLA-W0 $^{\dagger}$
& Pre-tokenized & -- & --
& 9,073.3 M& 17.23 GB & 1,633.5 ms& 86.1 \\

DriveLaW
& $768\!\times\!1344$ & 4 & 1
& 3,301.6 M& 6.89 GB & 1,167.0 ms & 88.6 \\

CoWorld-VLA
& $512\!\times\!1024$ & 1 & 1
& 5,372.9 M& 10.36 GB & 378.4 ms& 90.0 \\

Drive-JEPA
& $256\!\times\!512$ & 2 & 1
& 311.2 M& 1.83 GB & 73.3 ms& 87.8 \\

\textbf{MORE (Ours)}
& $256\!\times\!1024$ & 4 & 3
& 642.7 M& 6.18 GB & 110.0 ms & \textbf{92.5} \\
\bottomrule
\end{tabular}

\end{table*}

As shown in Tab.~\ref{tab:efficiency}, MORE achieves the highest EPDMS of 92.5 while maintaining an inference latency close to Drive-JEPA~\citep{drive-jepa}, which does not use an explicit future-state world model (110.0 VS. 73.3 ms). With comparable peak memory (6.18 vs. 6.89 GB), MORE requires only about 1/10 the inference latency of DriveLaW~\citep{drivelaw} (110.0 vs. 1,167.0 ms). These results show that motion-oriented registers provide an effective and efficient shared state interface for future world modeling and planning.

All measurements use one NVIDIA RTX 4090 with batch size 1, serial execution, warm-up, and synchronized CUDA events. Latency is the per-sample median, and memory is peak allocated GPU memory, including resident model weights. Disk I/O, dataset parsing, and PDM simulation are excluded. $^{\dagger}$ DriveVLA-W0~\citep{li2026drivevla} is evaluated from its released pre-tokenized multimodal interface. Thus, the number of raw camera views is unspecified, and image tokenization is excluded from the measured latency. 

\section{More Visualization}
\label{F}

\subsection{Additional Qualitative Trajectory Comparisons}

\begin{figure}[h]
\centering
\includegraphics[scale=0.243]{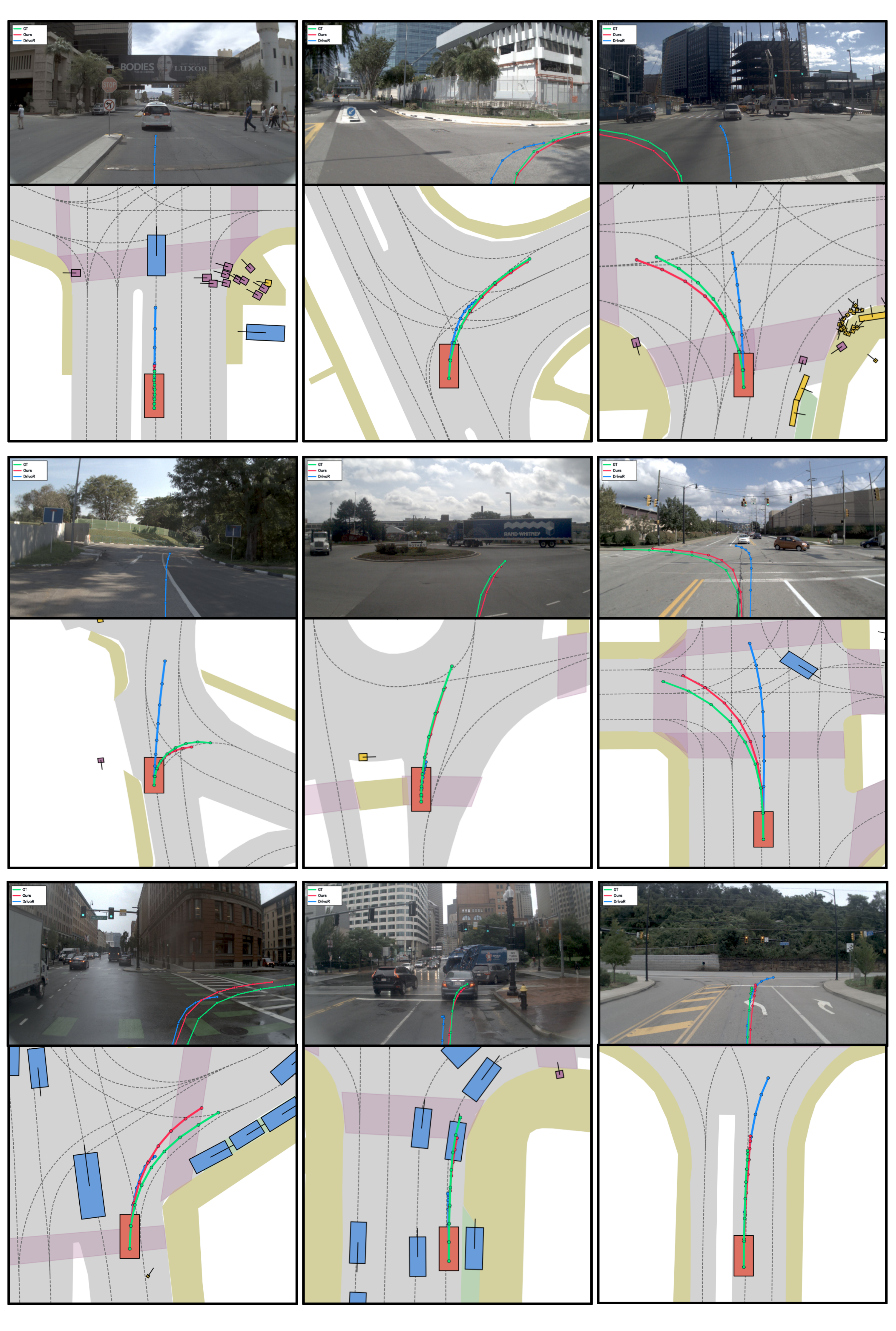}
\captionsetup{skip=1pt}
\caption{\textbf{Additional qualitative comparison.} 
Trajectories:
  \protect\tikz[baseline=-0.5ex]{\protect\shade[left color=green!70!black, right color=green!50!yellow] (0,-0.05) rectangle (0.4,0.05); \protect\fill[green!70!black] (0.2,0) circle (2.5pt);} GT,
  \protect\tikz[baseline=-0.5ex]{\protect\shade[left color=red!85!black, right color=red!65!white](0,-0.05) rectangle (0.4,0.05);\protect\fill[red] (0.2,0) circle (2.5pt);} MORE (Ours),
  \protect\tikz[baseline=-0.5ex]{\protect\shade[left color=cyan!80!blue, right color=cyan!50!blue] (0,-0.05) rectangle (0.4,0.05); \protect\fill[cyan!80!blue] (0.2,0) circle (2.5pt);} DrivoR.
}
\label{vis2}
\end{figure}

Fig.~\ref{vis2} presents additional comparisons across diverse straight and turning scenarios. Each example shows the front-camera view (top) and BEV visualization (bottom). MORE closely follows the GT travel direction and curvature, including at complex intersections with surrounding traffic. In contrast, DrivoR exhibits larger lateral deviations or continues straight in several turning cases. Together with Fig.~\ref{vis1}, these results indicate that motion-oriented registers retain planning-relevant motion and scene context for accurate trajectory generation.

\subsection{Register Similarity Analysis}
\label{sec:register_similarity}

\begin{figure}[H]
\centering
\includegraphics[width=\linewidth]{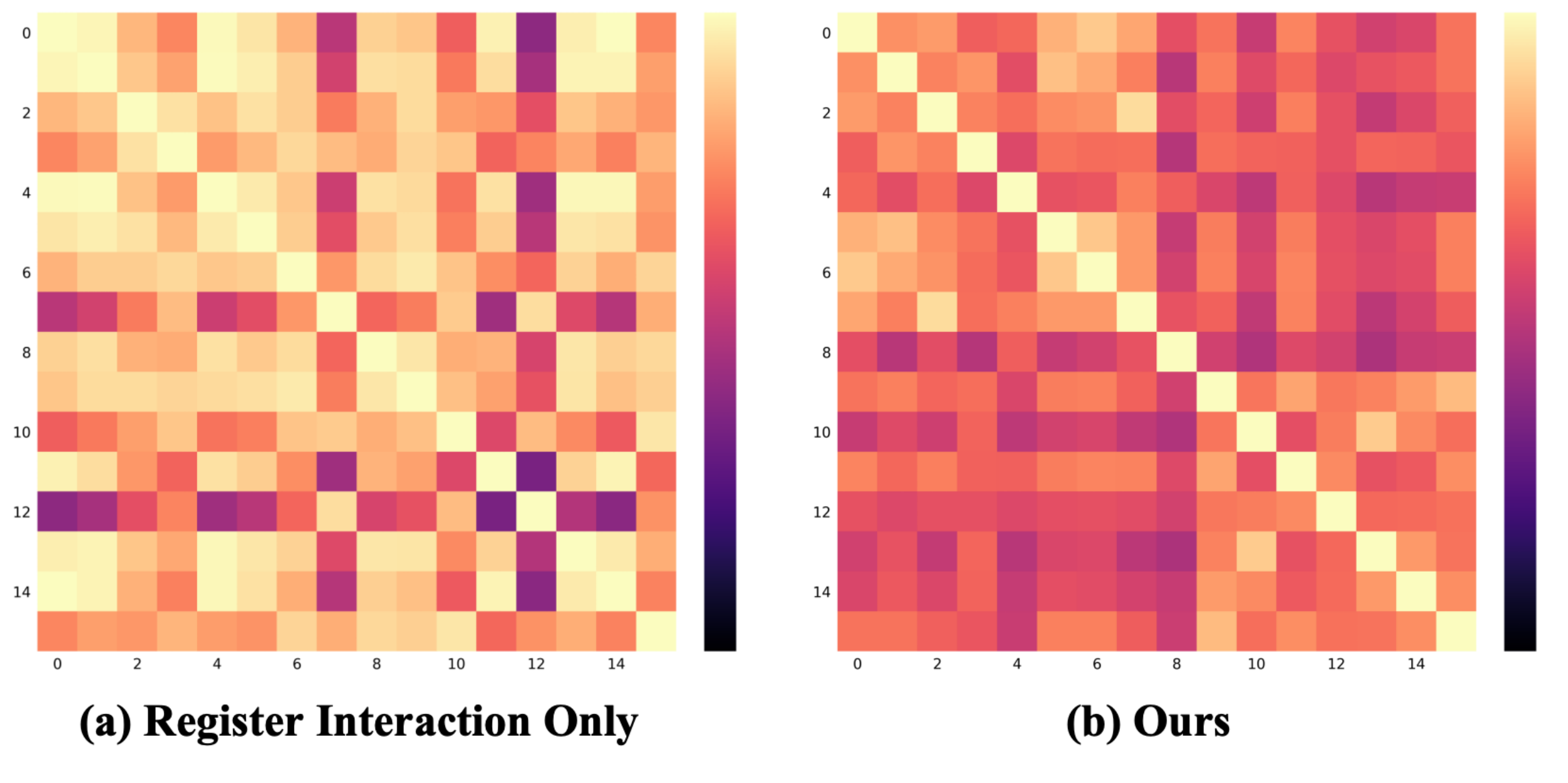}
\captionsetup{skip=1pt}
\caption{\textbf{Pairwise similarity of pretrained registers.}
Register similarities within and across temporal chunks are shown for
\textbf{(a) configuration \#4 in Tab.~\ref{tab:ablation1}, which enables register--patch interaction while retaining patch-based prediction, and
(b) MORE. }Their mean similarities are 0.793 and 0.653, respectively.}
\label{fig:register_similarity}
\end{figure}

To examine how our pretraining design shapes the register space beyond merely introducing register tokens, we compare the pretrained representations of configuration \#4 in Tab.~\ref{tab:ablation1} and MORE. As shown in Fig.~\ref{fig:register_similarity}, configuration \#4 exhibits broadly high similarity within and across temporal chunks, with a mean of 0.793, indicating substantial redundancy among registers. MORE reduces the mean similarity to 0.653 and produces more differentiated patterns across register slots and temporal chunks. These results show that our register pretraining design encourages complementary rather than highly overlapping representations, yielding a more expressive compact state interface for future world modeling and planning.

\subsection{Failure Case Analysis}
\label{sec:failure_cases}

\begin{figure}[H]
\centering
\includegraphics[width=\linewidth]{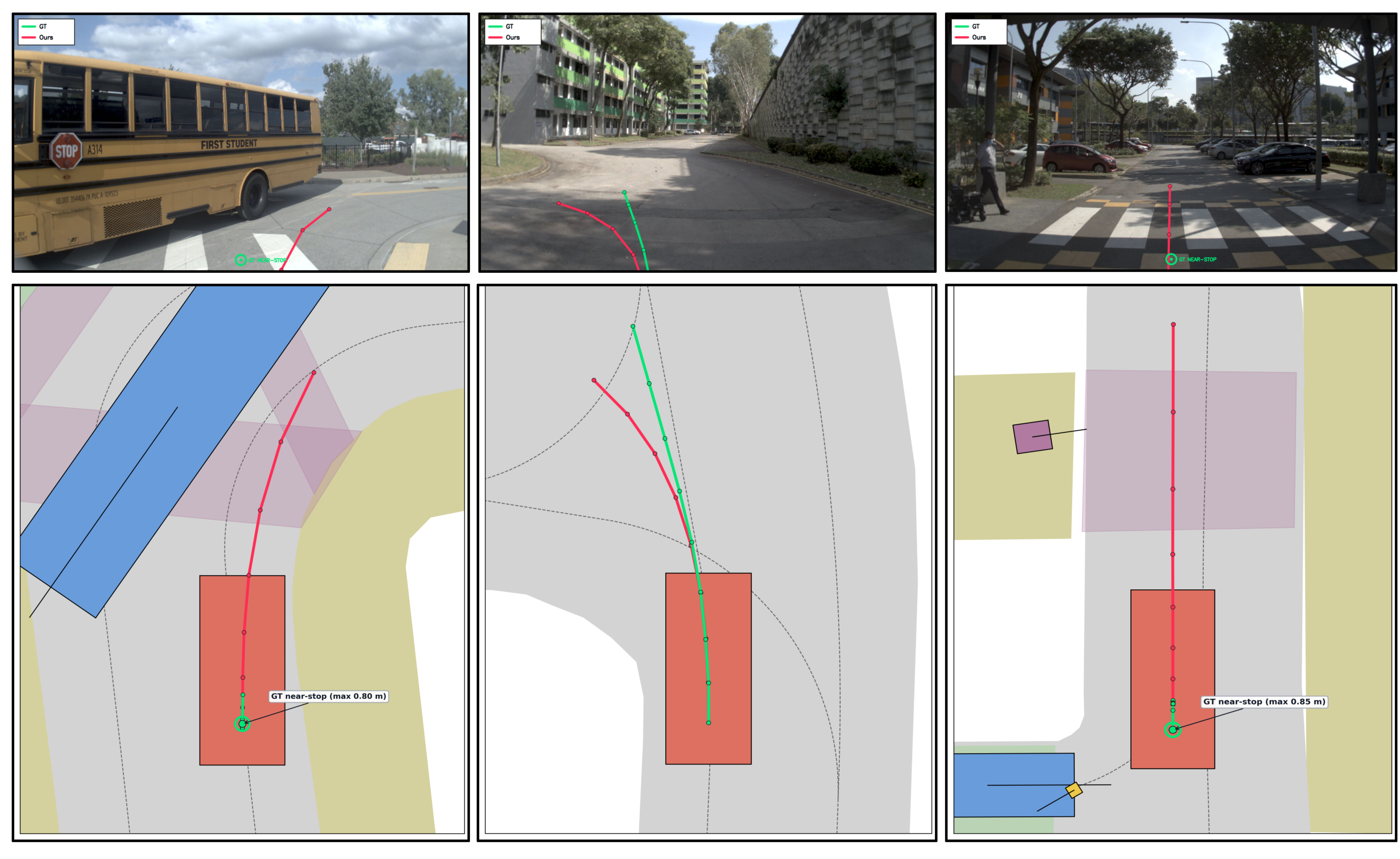}
\captionsetup{skip=1pt}
\caption{\textbf{Representative failure cases.}
Trajectories:
  \protect\tikz[baseline=-0.5ex]{\protect\shade[left color=green!70!black, right color=green!50!yellow] (0,-0.05) rectangle (0.4,0.05); \protect\fill[green!70!black] (0.2,0) circle (2.5pt);} GT,
  \protect\tikz[baseline=-0.5ex]{\protect\shade[left color=red!85!black, right color=red!65!white](0,-0.05) rectangle (0.4,0.05);\protect\fill[red] (0.2,0) circle (2.5pt);} MORE (Ours).
}
\label{fig:failure_cases}
\end{figure}

Fig.~\ref{fig:failure_cases} presents representative failure cases. In the two near-stop scenarios, the GT remains within approximately 0.85\,m of its initial position, whereas MORE advances by 9.9--11.7\,m. In the left example, the school bus severely occludes the road and potential cross traffic, and in the right example, the pedestrian crossing and surrounding agents provide ambiguous cues about the required stop duration. These errors are consistent with a common challenge in vision-based planning: under partial observability, a short observation window may admit multiple plausible future behaviors. The center example is substantially milder, with an average displacement error (ADE) of 0.46\,m and a final displacement error (FDE) of 1.80\,m, and occurs on an unmarked road with branch-like geometry and weak lane guidance. Overall, the failures are concentrated in scenes where the duration of a stop or the intended route continuation is difficult to infer from the available observations. Longer temporal context and uncertainty-aware future prediction may improve robustness in such cases.

\end{document}